\documentclass[conference,oneside]{IEEEtran}
\IEEEoverridecommandlockouts
\usepackage{cite}
\usepackage{amsmath,amssymb,amsfonts}
\usepackage{algorithmic}
\usepackage{graphicx}
\usepackage{textcomp}
\usepackage{xcolor}
\usepackage{float}
\usepackage{tabularx}
\usepackage[english]{babel}
\usepackage{enumitem}
\usepackage{tabularx}
\usepackage[
  top=0.75in,
  bottom=1.05in,
  left=0.75in,
  right=0.75in
]{geometry}
\renewcommand{\arraystretch}{1.2}
\newlist{steps}{enumerate}{1}
\setlist[steps, 1]{label = Step \arabic*:}
\def\BibTeX{{\rm B\kern-.05em{\sc i\kern-.025em b}\kern-.08em
    T\kern-.1667em\lower.7ex\hbox{E}\kern-.125emX}}
\begin{document}

\setlength{\textfloatsep}{2pt plus 1pt minus 2pt}
\setlength{\floatsep}{2pt plus 1pt minus 2pt}
\setlength{\intextsep}{2pt plus 1pt minus 2pt}

\setlength{\abovecaptionskip}{1pt}
\setlength{\belowcaptionskip}{1pt}

\setlength{\tabcolsep}{3pt}
\renewcommand{\arraystretch}{1}

\setlist[itemize]{topsep=2pt, itemsep=1pt, parsep=0pt, partopsep=0pt}
\setlist[enumerate]{topsep=2pt, itemsep=1pt, parsep=0pt, partopsep=0pt}

\setlength{\abovedisplayskip}{4pt}
\setlength{\belowdisplayskip}{4pt}
\setlength{\abovedisplayshortskip}{3pt}
\setlength{\belowdisplayshortskip}{3pt}

\title{Leveraging Multi-Agent System (MAS) and Fine-Tuned Small Language Models (SLMs) for Automated Telecom Network Troubleshooting}

\author{
Chenhua Shi
\quad Bhavika Jalli
\quad Gregor Macdonald
\quad John Zou
\quad Wanlu Lei
\quad Mridul Jain
\quad Joji Philip 
\\[0.25em]
\texttt{Ericsson} \\[0.25em]
}

\maketitle

\begin{abstract}

Telecom networks are rapidly increasing in scale and complexity, making management, operation, and optimization increasingly challenging. Although Artificial Intelligence (AI) has been applied to various telecom tasks, existing approaches are often limited in scope, require extensive labeled data, and struggle to generalize across heterogeneous deployments. Consequently, network troubleshooting still relies heavily on Subject Matter Experts (SMEs) to manually correlate multiple data sources and determine root causes and corrective actions. In this paper, we propose a Multi-Agent System (MAS) that leverages an agentic workflow in which Large Language Models (LLMs) coordinate specialized tools to enable automated network troubleshooting. Upon fault detection by AI/ML-based monitoring systems, the framework dynamically activates multiple agents—including an orchestrator, solution planner, executor, data retriever, and root-cause analyzer—to diagnose issues and recommend remediation strategies in near real time. A key component is the solution planner, which generates executable remediation plans grounded in internal operational documentation. To enable this capability, we fine-tune a Small Language Model (SLM) using proprietary troubleshooting documents to produce domain-specific solution plans. Experimental results demonstrate that the proposed framework significantly improves troubleshooting automation and efficiency across both Radio Access Network (RAN) and Core network domains.

\end{abstract}

\section{Introduction}
Telecommunication networks are evolving into highly dynamic and heterogeneous environments spanning multiple standards, vendors, and deployment scenarios~\cite{telecom_fm_trends}. This growing complexity makes troubleshooting particularly challenging, as it requires correlating diverse data sources such as performance metrics, configurations, alarms, and logs. Although AI/ML algorithms have been integrated into fault detection and troubleshooting, much of the diagnostic and remediation process is still performed manually by SMEs. This results in slow, resource-intensive, and difficult-to-scale operations. Consequently, autonomous monitoring and troubleshooting is becoming essential to reduce reliance on human expertise and improve operational efficiency.

Recent advances in Generative AI~\cite{generative_ai} and Foundation Models~\cite{foundationmodels}—particularly LLMs—have opened new opportunities for intelligent network automation. LLM-driven agentic systems, often combined with Retrieval-Augmented Generation (RAG)~\cite{rag}~\cite{ragsurvey}, have demonstrated strong reasoning and orchestration capabilities across multiple domains. Applied to telecom, such systems enable dynamic workflows in which specialized agents (e.g., solution planners, data retrievers, and root-cause analyzers) collaborate under the coordination of an LLM to perform complex troubleshooting tasks in a continuous REACT-style loop (Reasoning, Execution, and Action)~\cite{react}. However, practical deployment of these systems faces several challenges: (i) high operational costs associated with external LLM providers, (ii) data privacy risks when handling sensitive network information, and (iii) substantial capital expenditures (CapEx) required to host and deploy large models within operator environments.

To address these limitations, there is growing interest in SLMs as lightweight, domain-specialized alternatives to LLMs. SLMs are the future of agentic AI~\cite{slm_future_agentic_ai} since they can offer sufficient reasoning capacity for many agentic workflows while being more economical, privacy-preserving, and adaptable after fine-tuning. Nevertheless, effective use of SLMs requires high-quality domain datasets, efficient reinforcement fine-tuning (RFT) pipelines and robust evaluation criteria to ensure accuracy and reliability.

In this work, we propose a MAS with Human-in-the-Loop (HITL) for automated network troubleshooting that combines the general reasoning capabilities of an LLM with the efficiency of a fine-tuned SLM. The framework employs an LLM as the orchestration “brain” to coordinate agents such as a solution planner, executor, data retriever, and root-cause analyzer. A key innovation is the fine-tuned SLM solution planner, trained on internal troubleshooting documents, which generates domain-grounded remediation strategies tailored to RAN and Core networks. Experimental results demonstrate that this architecture significantly reduces troubleshooting time, alleviates SME workload, and enhances automation efficiency across heterogeneous deployment scenarios.

\section{Related Work}

Network troubleshooting is a critical operation in telecommunications and has been an active research area for decades, with prior surveys and taxonomies documenting diverse failure modes, diagnostic techniques, and persistent challenges \cite{network_troubleshooting_survey}. Traditional troubleshooting approaches typically combine rule-based logic, active probing, and expert-driven procedures. While effective in specific contexts, these methods face difficulties in scaling to large networks, handling heterogeneous telemetry, and reducing reliance on deep domain expertise \cite{network_troubleshooting_survey}. More recently, data-driven approaches such as anomaly detection and predictive maintenance have shown promise in reducing manual effort. However, these methods often require high-quality labeled datasets, are sensitive to distribution shifts across deployments, and frequently lack explainability as well as true end-to-end automation \cite{singh_predictive_troubleshooting}.

Recent advances in GenAI have motivated research into applying large pretrained models to telecom tasks~\cite{telecom_fm_trends}~\cite{generative_ai_survey}~\cite{large_language_models_for_telecom}. These works highlights that combining LLMs with RAG can yield strong reasoning capabilities and knowledge-grounded responses, enabling richer natural-language interfaces and more flexible decision-making compared to classical ML pipelines. Despite these benefits, several practical challenges arise when deploying LLMs- and RAG-based solutions in operational telecom environments. These include:  
\begin{enumerate}[label=(\roman*)]  
    \item High cost and latency associated with large external LLMs. 
    \item Privacy and regulatory restrictions on exposing sensitive network telemetry to third-party services. 
    \item Difficulties in reliably grounding model outputs on noisy and heterogeneous network data.  
    \item Lack of end-to-end orchestration that integrates detection, diagnosis, planning, and execution.
\end{enumerate}  
These limitations underscore the need for more efficient, domain-adapted approaches, such as fine-tuned SLMs that can deliver accurate and context-aware troubleshooting within telecom constraints. Our work builds on these directions by (i) implementing a MAS with HITL that integrates LLM-based orchestration with specialized agents for retrieval, root-cause analysis, execution, and display, and (ii) fine-tuning a SLM (via SFT + RFT) to serve as a domain-grounded solution planner that is cost-effective and privacy-aware. Compared to earlier surveys and proposals, our contribution is to demonstrate an end-to-end, evaluated pipeline that addresses both the operational concerns (cost, privacy, deployment) and the technical challenges (grounding, consistency, actionability) of automated telecom troubleshooting.

\section{METHODOLOGY}

Our approach combines a MAS with a fine-tuned SLM on solution planner to enable fully automated network troubleshooting. The MAS leverages an LLM-driven agentic workflow to orchestrate specialized agents—including orchestrator, solution planner, data retriever, executor, and root-cause analyzer, dashboard display—within a ReAct-style loop for fault detection, analysis, and remediation. To ensure accurate and domain-grounded solution planning, we fine-tune the SLM through SFT and RFT using the Transformers Reinforcement Learning (TRL) framework. This design achieves scalable, efficient, and privacy-preserving troubleshooting across heterogeneous telecom networks.

\subsection{Multi Agent System for Automated Network Troubleshooting}

\begin{figure}[htbp!]
\centering
\includegraphics[width=0.9\linewidth]{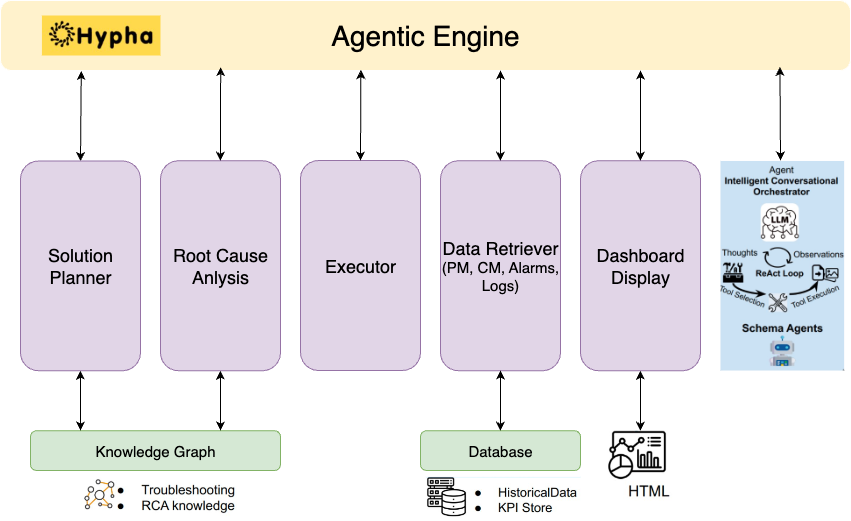}
\caption{\label{fig:MAS}The architecture diagram for Multi Agent System (MAS) for Automated Network Troubleshooting.}
\end{figure}

We present a MAS for automated network troubleshooting, as illustrated in Fig.~\ref{fig:MAS}. The MAS is built on top of the Hypha framework~\cite{hypha_framework}, a real-time application framework designed for agentic workflows. Hypha extends FastAPI by enabling Remote Procedure Calls (RPC) over WebSockets and HTTP, while providing features tailored for agent-based implementations. These include API annotations for schema agents, seamless integration of LLM agents with local or remote tools, and automatic service discovery of agents and tools. To orchestrate these components, we leverage Schema Agent~\cite{bioimageio_chatbot}, an LLM-powered workflow engine that supports error recovery through the ReAct loop, as well as API calls and code generation/execution. Together, Hypha, Schema Agent, and the ReAct Loop enable seamless and efficient agent-to-agent (A2A) interactions, forming the foundation of our automated troubleshooting framework. This design reduces inter-agent communication overhead and improves failure handling within the MAS.

Within the MAS, we developed six specialized agents, each responsible for distinct tasks. The intelligent conversational orchestrator serves as the central controller, interpreting user intents or detected network faults and delegating tasks to the appropriate agents. The solution planner agent is one of the most critical components, tasked with generating accurate, reliable, and stepwise troubleshooting procedures. To strengthen its reasoning, it is integrated with an external knowledge graph built from telecom troubleshooting documents—including performance metrics (PM), fault alarms (FM), and configuration management (CM) data. We employ HippoRAG~\cite{hipporag} as the retrieval backbone, which consistently outperforms standard RAG and GraphRAG in our internal evaluations. HippoRAG enhances grounding by integrating LLM-extracted triples with phrase- and passage-level knowledge graphs, while employing a personalized PageRank algorithm to prioritize the most contextually relevant information.

The Root Cause Analysis (RCA) agent is similarly equipped with a domain knowledge graph and is responsible for correlating detected faults across multiple data sources, including PM, CM, FM, logs, and RCA-specific knowledge. For PM data, we additionally extract causal intervention sequences~\cite{sdg} starting from the onset of abnormal failures. Based on this information, the agent identifies the underlying root cause and provides corresponding remediation strategies. The executor agent refines the troubleshooting plans generated by the solution planner, augmenting them with detailed parameters such as node identifiers, time intervals, and metric specifications required for data retrieval or analysis. Most importantly, we incorporate HITL feedback to validate generated executable plans. Once approved, the executor proceeds with implementation. Approximately 80\% of the workflow operates autonomously, delivering significant efficiency gains. Around 15\% consists of approval gates at critical stages to ensure no automated action impacts live network traffic without human consent. The remaining 5\% involves human decisions on business impact and prioritization, where experienced network operations professionals remain essential.

The data retriever agent provides multiple tools for accessing diverse databases, including a PM retriever for RAN counters stored in PostgreSQL, a PM retriever for Core counters in VictoriaMetrics, and dedicated retrieval tools for alarms and logs stored in OpenSearch. Finally, the dashboard display agent consolidates and visualizes the results in an interactive HTML interface, enabling SMEs to easily review and validate detected network faults.

\subsection{Fine-tuning on SLM for Solution Planner with TRL}

To adapt the SLM for solution planning, we adopt a two-stage fine-tuning strategy. First, we perform SFT as a warm-up phase, where the model is trained on labeled QA pairs to quickly learn domain context and response formatting. Next, we apply RFT, which guides model behavior through feedback and rewards rather than explicit labels. Here, we have applied Low-rank adaptation (LORA)~\cite{lora2021} and Group Relative Policy Optimization (GRPO)~\cite{deepseekmath} in RFT, which can effectively save memory and time during training. This stage strengthens the model’s reasoning ability, enabling it to produce accurate, domain-grounded troubleshooting steps. Training is implemented using the TRL library, which supports multi-GPU execution to accelerate the process.

\begin{figure}[htbp!]
\centering
\includegraphics[width=0.8\linewidth]{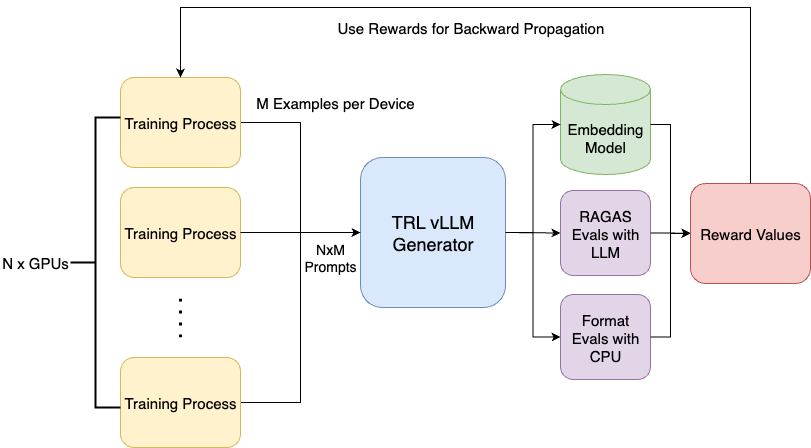}
\caption{\label{fig:TRL}RFT pipeline for fine-tuning a Small Language Model (SLM) as a solution planner using Transformers Reinforcement Learning (TRL) across multiple GPUs.}
\end{figure}

Figure~\ref{fig:TRL} illustrates the RFT pipeline under TRL. In a multi-GPU setup with
$N$ devices, each GPU processes a batch of
$M$ examples, collectively generating
$N$ x $M$ prompts for the TRL-vLLM engine. vLLM~\cite{vllm}, an open source LLM serving framework integrated with TRL, enables fast response generation, which is then passed to the reward evaluation modules. Reward signals are computed using a combination of hosted embedding models and LLMs for RAGAS~\cite{ragas}-based evaluations, while additional CPU-based checks (e.g., regex validation) ensure proper formatting. These reward values are backpropagated to update the model parameters. The training proceeds in a cyclical loop: the TRL-vLLM generator produces candidate responses, which are then evaluated for quality and grounding, followed by reward propagation to the training process. Due to backpropagation, memory consumption spikes during updates, after which the GPUs idle until the next batch of generated responses becomes available. This alternating cycle of generation, evaluation, and reward propagation continues throughout the entire RFT process.
 
\section{EXPERIMENTS}

We evaluate the proposed framework through two sets of experiments: (i) deployment of the MAS for automated troubleshooting across RAN and Core network domains, and (ii) RFT of the Solution Planner using a SLM. These experiments assess both the end-to-end troubleshooting performance and the effectiveness of fine-tuned models in generating domain-grounded remediation steps.

\subsection{AI Agents for Autonomous Network Operations}

We evaluate the proposed MAS framework for automated network troubleshooting across both RAN and Core network domains. In the RAN domain, the experiments focus primarily on power system–related issues, heartbeat failure alarm, and RAN capacity congestion. While in the core network, we focus on troubleshooting cases involving PDU session degradation. Meanwhile, we use GPT-4o mini in these experiments.

\subsubsection{Power System in RAN}

We index power system troubleshooting documents into HippoRAG during an offline phase, forming a knowledge graph used by the solution planner and RCA analyzer. When monitored KPIs exceed Service-Level Agreement (SLA) thresholds or alarms are triggered, the detection module automatically generates an intent prompt. For example, when an \textit{Input Power Failure} is detected for a \textit{ManagedElement} and \textit{FieldReplaceableUnit}, the orchestrator receives:

\begin{quote}
\textit{``Can you help me find Input Power Failure issues and top offenders in the last 15 minutes for triage?''}
\end{quote}

The conversational orchestrator invokes the solution planner to generate stepwise troubleshooting actions. An example plan includes:

\begin{enumerate}
    \item Identify correlated alarms with higher severity.
    \item Determine whether the issue affects FRUs.
    \item Monitor the \textit{FieldReplaceableUnit.pmPowerFailure} counter to detect persistent supply issues.
    \item Examine energy meter counters (e.g., \textit{pmVoltage}, \textit{pmCurrent}, and min/max values) to assess voltage and current stability.
    \item If unresolved, perform onsite inspection or escalation.
\end{enumerate}

The executor agent augments the plan with node identifiers and time intervals to produce executable actions, which are presented to the user for approval. Once approved, the executor triggers the data retriever agent to collect metrics from multiple sources in parallel. The RCA analyzer correlates the retrieved data with RCA knowledge and SME inputs to determine the root cause.

In this case, faults were localized to specific RRUs, indicating site-level power distribution issues rather than a full site outage. Anomalies in \textit{pmPowerFailure}, \textit{pmVoltage}, and \textit{pmCurrent1} confirmed isolated outages, likely caused by maintenance activities or brief interruptions. Finally, the dashboard display agent generates a network analysis report—including an executive summary, analysis results, actions taken, and recommended next steps—for SME validation.

\subsubsection{Heartbeat Failure Alarm in RAN}
The framework extends naturally to \textit{HeartbeatLost} alarms, where a managed element ceases sending periodic notifications to the SMO over the O1 interface. By indexing O1 interface and transport-layer documents into HippoRAG, the solution planner generates steps covering transport connectivity checks, O1 session validation (\textit{"presentSeverity": "CRITICAL", "specificProblem": "Heartbeat Failure"}), and NTP synchronisation verification. In a representative case, the RCA analyzer identifies NTP drift as the root cause, resolving the alarm without a site visit.

\subsubsection{RAN Capacity Congestion}
For congestion events triggered by sustained DL PRB utilisation above SLA thresholds, the framework retrieves \textit{pmRadioTechDlCapUsage}, \textit{pmRrcConnMax}, and neighbour cell load counters to identify over-conservative MLB parameters or underutilised carrier aggregation as root causes. In a representative case, adjusting the A5 handover threshold by 4~dB reduces peak PRB utilisation from 91\% to 67\% in the following ROP with no service impact.

\subsubsection{PDU Session Degradation in Core}
For degradation events triggered by abnormal drops in the monitored KPI \textit{pdu\_session}, the framework retrieves session metrics from VictoriaMetrics, including \textit{pdu\_session\_ipv4}, \textit{pdu\_session\_ipv6}, and \textit{subscriber\_count\_5g}, together with signaling counters such as \textit{pdu\_session\_create\_sm\_context\_resp}, \textit{comm\_n1n2\_msg\_transfer\_resp}, and \textit{session\_establishment\_resp\_acc\_rcvd}, to analyze traffic and signaling behavior. The framework further examines OpenSearch logs and alarms, including \textit{SingleHttpConnectionLost}, and performs configuration checks by verifying the SMF service status from the NRF registry. In a representative case, the RCA analyzer identifies a suspended NRF service as the primary cause of the degradation. The failure disrupts SMF registration and signaling exchanges, leading to session establishment failures and reduced PDU session counts. Correlated increases in signaling responses further confirm the anomaly. Restoring the NRF service reestablishes normal control-plane interactions, stabilizes session management, and prevents further PDU session drops.

\begin{figure}[htbp!]
\centering
\includegraphics[width=0.6\linewidth]{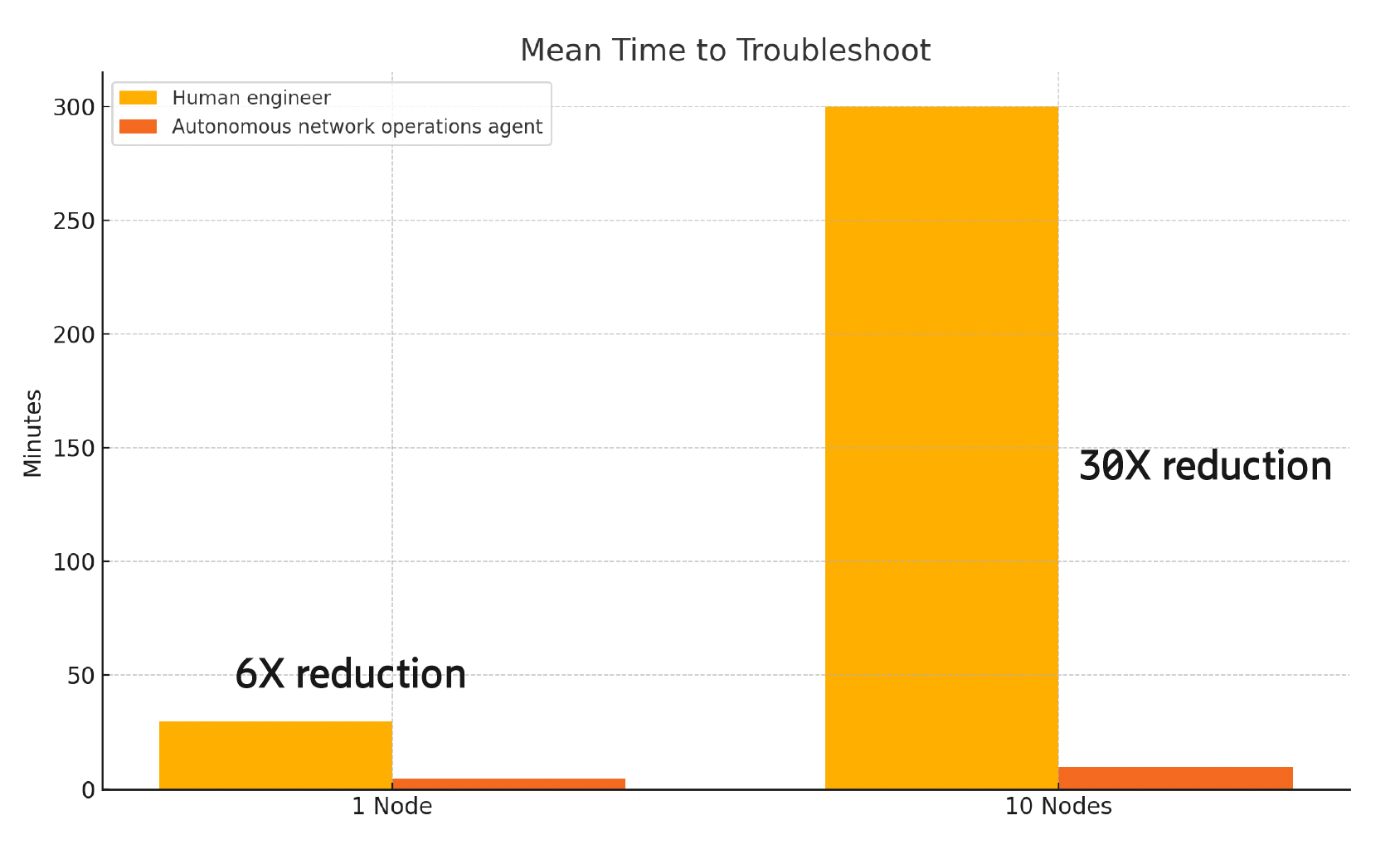}
\caption{\label{fig:TroubleshootTime}Autonomous Network Operations Agent Mean Time to Troubleshoot.}
\end{figure}

\begin{figure}[htbp!]
\centering
\includegraphics[width=0.6\linewidth]{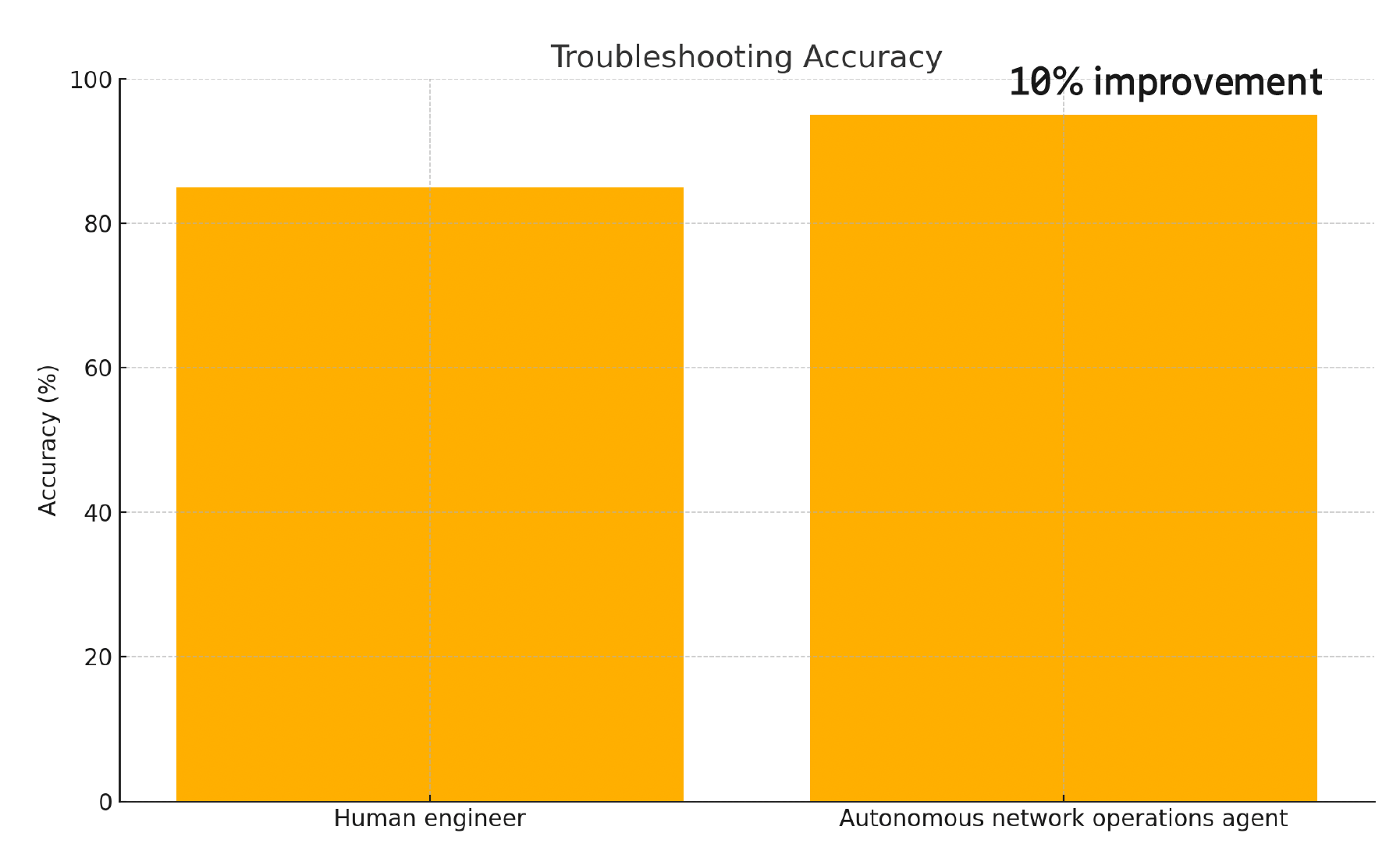}
\caption{\label{fig:TroubleshootAccuracy}Autonomous Network Operations Agent Troubleshooting Accuracy.}
\end{figure}

5) \textit{Autonomous Network Troubleshooting Benefits:}
Compared with human engineers in troubleshooting network faults, the autonomous network operations MAS demonstrates superior performance, achieving a six-fold reduction in mean troubleshooting time per node and a 10\% improvement in accuracy, as shown in Figures~\ref{fig:TroubleshootTime} and~\ref{fig:TroubleshootAccuracy}. As we can see, AI agents deliver quantifiable gains: faster troubleshooting, lower operation costs, and better customer experience-turning network complexity into competitive advantage.

\subsection{Fine-tuning on SLM for Solution Planner}

As illustrated earlier, the solution planner serves as a pivotal agent, as it must generate accurate troubleshooting steps by leveraging documents across diverse data sources. However, practical deployment faces challenges such as the high costs of external LLMs with long context windows and data privacy restrictions. To address these issues, we fine-tune a SLM  for solution planning. Our experiments are conducted on a cluster equipped with seven NVIDIA RTX A6000 GPUs, each with 48 GB of VRAM. We utilize a telecom troubleshooting seed dataset along with an RFT dataset for fine-tuning~\cite{thinkless2025}, selecting \texttt{Unsloth/DeepSeek-R1-Qwen-3-8B} as the base model due to its reasoning capability, suitability for solution planning, and parameter efficiency under our GPU budget. Specifically, we allocate one GPU for vLLM serving the embedding model \texttt{Qwen/Qwen3-Embedding-0.6B}, one GPU for vLLM serving \texttt{Qwen/Qwen3-8B} with disabled thinking for RAGAS evaluation, and two GPUs for TRL vLLM serving with \texttt{Unsloth/DeepSeek-R1-Qwen-3-8B}. In addition, we employ ZeRO Stage-2 from DeepSpeed~\cite{zeroptimizer2019} to optimize memory usage during training.

\subsubsection{Improvements on the Context Length and Training Speed}


During the RFT phase, we adopt a large window chunk size because the top-3 retrieved documents are concatenated with the query as input, resulting in significantly higher memory usage than typical training. The extended context length increases both memory consumption and training time. We observe that HippoRAG retrieves full documents without internal chunking, even when only partial content is relevant. To address this, we integrate LangChain-based document chunking. While semantic chunking groups related passages, it produces non-uniform chunk sizes. In contrast, uniform chunking provides a consistent upper bound on context length and proves more effective for training. After applying uniform chunking, training efficiency improves substantially: the original 1000 training steps execute faster and converge more smoothly. Furthermore, chunking enables higher LoRA ranks, larger batch sizes, and more generations per step, improving convergence and overall training throughput. With uniform chunking, the full training process completes within three days, while resuming from a checkpoint requires only about 20 hours.

\subsubsection{Reward Design for RFT in Network Troubleshooting}

Reward functions are a critical component in GRPO-based policy optimization, as they guide the model to strengthen its reasoning capability through scoring of generated candidates. In our framework, we employ two categories of reward functions: (i) Format Rewards, and (ii) RAGAS-based Rewards customized for telecom troubleshooting. The format rewards include regex-based checks to incentivize the correct use of XML tags for answer formatting and step demarcation, thereby facilitating reliable parsing. The RAGAS-based rewards are adapted to telecom-specific requirements.

    \begin{itemize}
        \item Answer and Reasoning Completeness: Encourages the model to generate stepwise troubleshooting responses that incorporate all relevant PM counters, fault alarms, and configuration commands, thereby ensuring comprehensive coverage of network diagnostic factors.
        \item Answer and Reasoning Relevancy: Ensures that both the proposed actions and their underlying reasoning remain directly aligned with the specific troubleshooting query, avoiding extraneous or domain-irrelevant content.
        \item Answer and Reasoning Groundedness: Verifies that all referenced entities—such as counters, alarms, and network components—are explicitly supported by the retrieved context, thereby reducing hallucinations and improving factual reliability.
    \end{itemize}

\subsubsection{RFT Results Comparison on Different Models}

\begin{figure}[htbp!]
\centering
\includegraphics[width=0.9\linewidth]{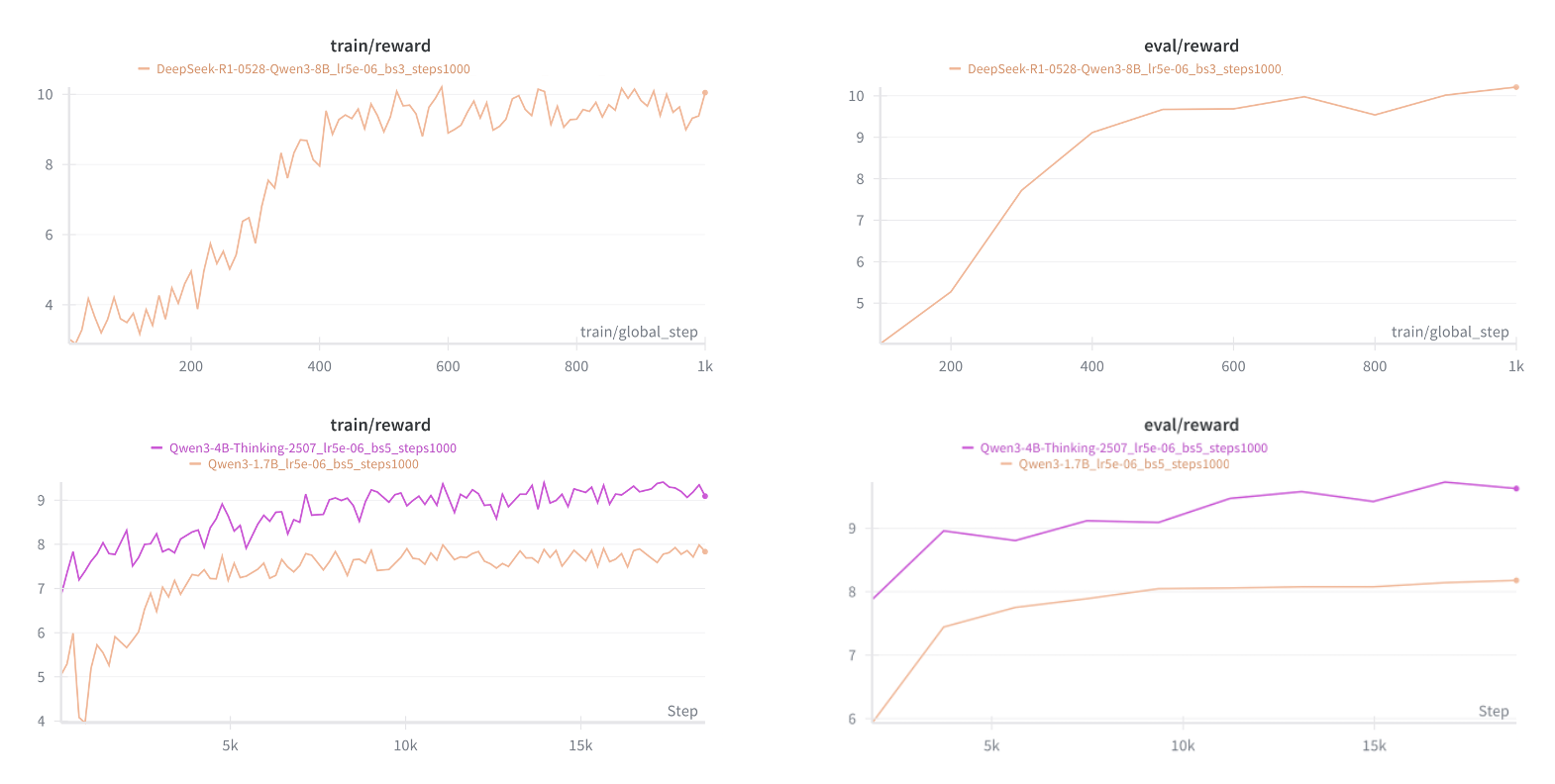}
\vspace{-1em}
\caption{\label{fig:TrainMean}Mean of Rewards on Training and Evaluation among Different Models.}
\end{figure}

\begin{figure}[htbp!]
\centering
\includegraphics[width=0.9\linewidth]{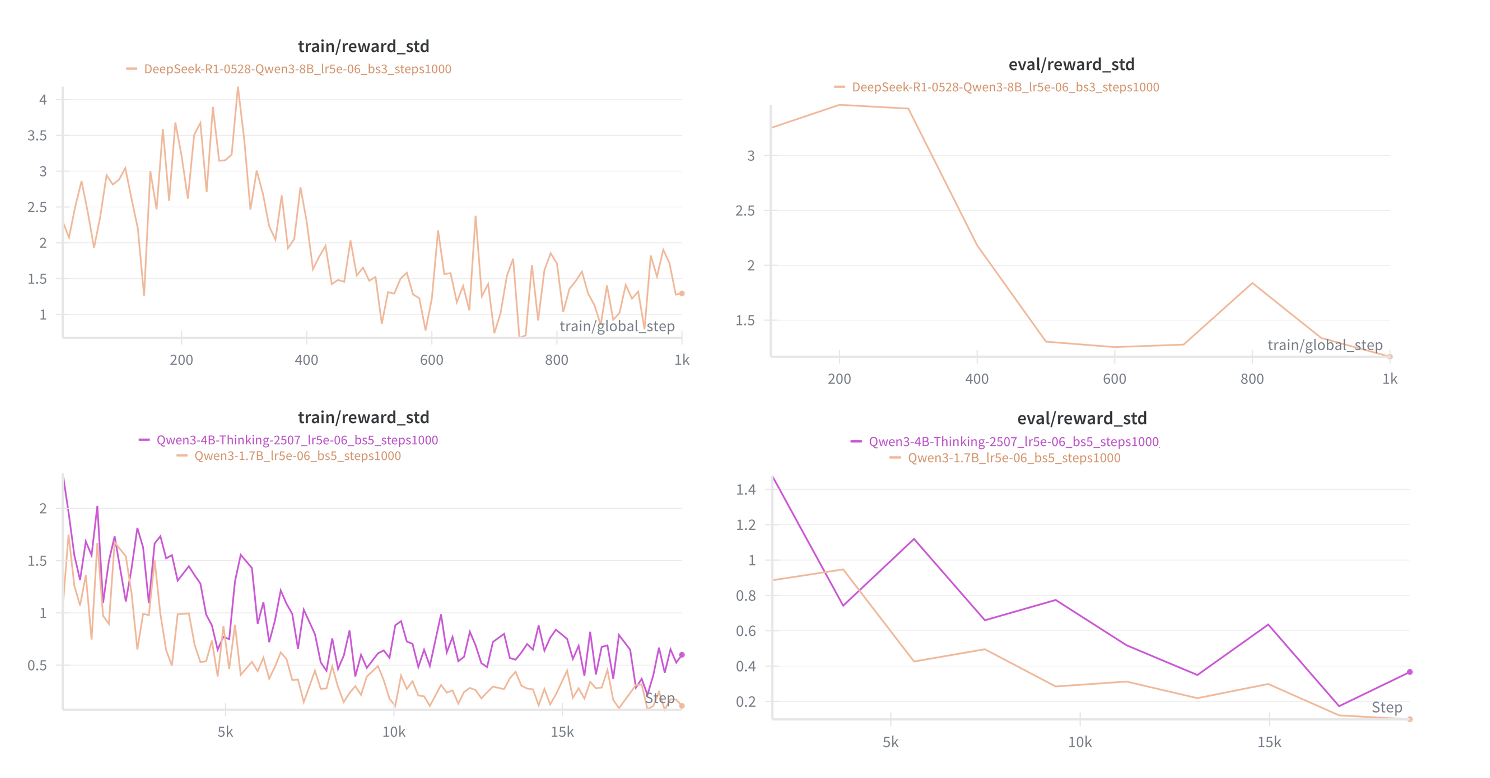}
\vspace{-1em}
\caption{\label{fig:TrainStd}Standard Deviation of Rewards on Training and Evaluation among Different Models.}
\end{figure}

We perform RFT on three models of varying sizes: \texttt{DeepSeek-R1-Qwen-3-8B}, \texttt{Qwen3-4B}, and \texttt{Qwen3-1.7B}. As shown in Figures~\ref{fig:TrainMean} and~\ref{fig:TrainStd}, the 8B model achieves the best performance, reaching peak rewards near 10. The 4B and 1.7B models show modest improvements but consistently lower rewards. A notable observation is the reduction in reward standard deviation across all models, with the fine-tuned 8B model exhibiting the most stable performance. The lower variance indicates more consistent outputs across runs, and stability further improves as fine-tuning progresses, producing higher rewards with reduced variability.

\begin{figure}[htbp!]
\centering
\includegraphics[width=1\linewidth]{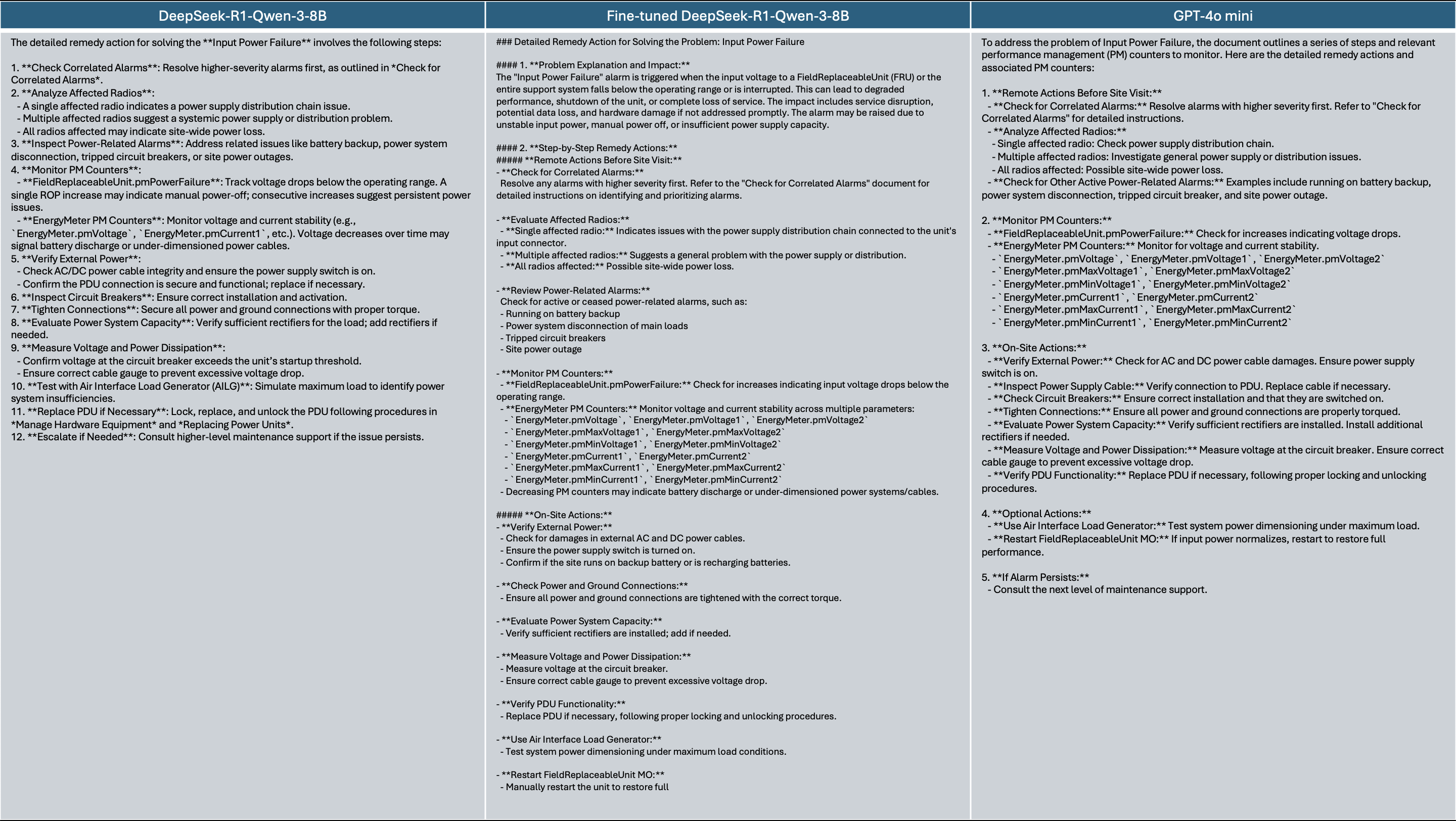}
\caption{\label{fig:RFTCompare}Comparison of Responses from Base Model, Fine-Tuned Model, and GPT-4o mini.}
\end{figure}

Table~\ref{tab:ModelsComparsion} shows that the fine-tuned 8B model achieves higher rewards and significantly lower variance than the base model. As illustrated in Figure~\ref{fig:RFTCompare}, for the \textit{Input Power Failure} alarm, the fine-tuned model generates solutions comparable to GPT-4o-mini, including detailed counter-level troubleshooting steps. In contrast, the base model produces only generic responses lacking actionable guidance.

\begin{table}[htbp!]
\centering
\caption{Comparison between Original Model and Fine-tuned Model on DeepSeek-R1-Qwen-3-8B}
\begin{tabular}{|m{2.6cm}|
                >{\centering\arraybackslash}m{1.5cm}|
                >{\centering\arraybackslash}m{1.5cm}|
                >{\centering\arraybackslash}m{1.5cm}|}
\hline
 \textbf{Metrics} & \textbf{Original Model} & \textbf{Finetuned Model} \\
\hline
RAGAS Rewards & 3.44 & 5.19 \\
\hline
Format Rewards & 2.52 & 5.31 \\
\hline
Total Rewards & 5.96 & 10.51 \\
\hline
Standard Deviation & 2.17 & 0.38 \\
\hline
\end{tabular}
\label{tab:ModelsComparsion}
\end{table}

\subsubsection{Ablation Study}
Isolate the contribution of each technical choice, we conduct two ablation experiments.

\paragraph{Retrieval backend} We compare solution plan quality under Standard RAG, GraphRAG, and HippoRAG using identical prompts and documents. HippoRAG's personalised PageRank retrieves fault-specific PM counters and alarm names with higher precision, reducing hallucinated entities in generated plans. Total RAGAS rewards are 3.61 (Standard RAG), 4.08 (GraphRAG), and 5.19 (HippoRAG), with HippoRAG also yielding the lowest variance (std. dev. 0.38 vs. 1.42--1.89).

\paragraph{Fine-tuning stages} Table~\ref{tab:AblationFT} extends Table~\ref{tab:ModelsComparsion} with an SFT-only baseline. SFT alone increases the total reward from 5.96 to 8.31 and reduces variance, but generated plans often omit specific counter names. RFT, guided by Answer Completeness and Groundedness rewards, further improves the total reward by 2.20 to 10.51 and reduces the standard deviation to 0.38, producing counter-level detail comparable to GPT-4o-mini.

\begin{table}[htbp!]
\centering
\caption{Ablation of fine-tuning stages on DeepSeek-R1-Qwen-3-8B}
\begin{tabular}{|m{2.6cm}|
                >{\centering\arraybackslash}m{1.5cm}|
                >{\centering\arraybackslash}m{1.5cm}|
                >{\centering\arraybackslash}m{1.5cm}|}
\hline
\textbf{Metrics} & \textbf{Base} & \textbf{SFT Only} & \textbf{SFT+RFT} \\
\hline
RAGAS Rewards   & 3.44 & 4.21 & 5.19 \\
\hline
Format Rewards  & 2.52 & 4.10 & 5.31 \\
\hline
Total Rewards   & 5.96 & 8.31 & 10.51 \\
\hline
Standard Deviation  & 2.17 & 1.02 & 0.38 \\
\hline
\end{tabular}
\label{tab:AblationFT}
\end{table}

\section{CONCLUSION}

In this work, we presented a MAS for automated network troubleshooting that integrates LLM-based orchestration with a fine-tuned SLM solution planner. The framework coordinates specialized agents for orchestration, planning, data retrieval, execution, root-cause analysis, and dashboard reporting. Experimental results across RAN and Core domains show that the proposed approach reduces troubleshooting mean time by up to 6× per node and improves accuracy by 10\% compared to human-driven processes. In addition, reinforcement fine-tuning with domain-specific reward functions enables the SLM planner to generate precise, context-grounded troubleshooting strategies while reducing reliance on costly external LLMs. Overall, the results demonstrate that combining LLM orchestration with fine-tuned SLMs provides a scalable, efficient, and privacy-preserving solution for telecom network operations automation.

\vspace{12pt}

\end{document}